\documentclass{article}

\usepackage[T1]{fontenc}

\usepackage{microtype}
\PassOptionsToPackage{hyphens}{url} %
\usepackage{graphicx}
\usepackage{amsmath,amssymb}
\usepackage{booktabs} %
\usepackage{placeins} %
\usepackage{flafter} %

\usepackage{xcolor}
\definecolor{soLink}{RGB}{0,60,130}
\usepackage[colorlinks=true,linkcolor=soLink,citecolor=soLink,urlcolor=soLink]{hyperref}

\usepackage{xspace}
\newcommand{\sysname}{the qualification labeling system\xspace}
\newcommand{\Sysname}{The qualification labeling system\xspace}

\newif\ifsoblind
\soblindfalse

\newif\ifsodraft
\sodraftfalse
\definecolor{soOwner}{RGB}{160,60,0}
\newcommand{\sectionowner}[2][]{\ifsodraft\par
  {\raggedright\noindent\color{soOwner}\sffamily\footnotesize[Contributors:~#2%
   \if\relax\detokenize{#1}\relax\else,~#1\fi]\par}\nobreak\vspace{-\parskip}\noindent\fi\ignorespaces}

\IfFileExists{mlsys2026.sty}{\usepackage[accepted]{mlsys2026}}{%
  \IfFileExists{mlsys2025.sty}{\usepackage[accepted]{mlsys2025}}{%
    \usepackage{fallback}}}
\ifdefined\hypersetup\hypersetup{pdfsubject={MLSys 2027}}\fi

\makeatletter
\ifdefined\isaccepted\else
  \renewcommand{\mlsysauthor}[2]{%
    \ifdefined\@icmlfirsttime\else
      \gdef\@icmlfirsttime{1}%
      \mbox{\bf Anonymous Author(s)}\addtofullauthorlist{Anonymous Author(s)}%
    \fi\ignorespaces}%
  \renewcommand{\printAffiliationsAndNotice}[1]{%
    {\let\thefootnote\relax\footnotetext{\hspace*{-\footnotesep}%
      Anonymous Institution. Correspondence to: Anonymous Author(s) \textless{}anon.email@domain.com\textgreater{}.
      \ \\ \Notice@String}}}
\fi
\makeatother

\makeatletter\providecommand{\mlsys@appearing}{}\renewcommand{\mlsys@appearing}{}\makeatother

\providecommand{\href}[2]{#2}

\mlsystitlerunning{From Generation to Scoring}

\begin{document}
\raggedbottom
\setcounter{topnumber}{3}
\setcounter{bottomnumber}{2}
\setcounter{totalnumber}{4}
\renewcommand{\topfraction}{0.9}
\renewcommand{\bottomfraction}{0.6}
\renewcommand{\textfraction}{0.07}
\renewcommand{\floatpagefraction}{0.8} %

\twocolumn[
\mlsystitle{From Generation to Scoring: Scaling LLM-Based Semantic Labeling at LinkedIn}

\mlsyssetsymbol{equal}{*}

\begin{mlsysauthorlist}
\mlsysauthor{Jian Sheng}{li,equal}
\mlsysauthor{Sriram Vasudevan}{li,equal}
\mlsysauthor{Sundara Raman Ramachandran}{li,equal}
\mlsysauthor{Yang Hu}{li,equal}
\mlsysauthor{Junyu Zhou}{li,equal}
\\
\mlsysauthor{Xiaojing Ma}{li,equal}
\mlsysauthor{Haolan Qin}{li}
\mlsysauthor{Tiffany Zhou}{li}
\mlsysauthor{Vedant Jhaveri}{li}
\mlsysauthor{Ashwini Bhingare}{li}
\mlsysauthor{Chinmay Kini}{li}
\\
\mlsysauthor{Qing Lan}{li}
\mlsysauthor{Xiaoqing Wang}{li}
\mlsysauthor{Suleyman Cetintas}{li}
\mlsysauthor{Fedor Borisyuk}{li}
\end{mlsysauthorlist}

\mlsysaffiliation{li}{LinkedIn}

\mlsyscorrespondingauthor{Jian Sheng}{jsheng@linkedin.com}

\mlsyskeywords{Machine Learning, MLSys, LLM Inference, Semantic Relevance}

\vskip 0.3in

\begin{abstract}
\noindent For efficiency, industrial large language model (LLM) labelers often pack many candidate labels into one prompt and generate a \textsc{yes}/\textsc{no} answer for each, even when each decision depends only on the entity and one candidate.
Generating these independent decisions as one answer list couples them, spends decoding on the list, and relies on the list being well-formed.
In this paper, we present an end-to-end recipe that reformulates the labeler as \emph{semantic scoring}, fine-tuning it on single-candidate prompts to return each decision as one answer-token score from a prefill-only pass, and we adapt Multiple-Item Scoring (MIS) to score hundreds of candidates in one sequence with attention masked between them.
Deployed for profile qualification labeling at LinkedIn, semantic scoring changes at most 0.01\% of decisions when candidates are reordered and matches the agreement with reference labels that packed generation reaches only with about $4.7\times$ as many training pairs, and the MIS-based stack raises throughput over the deployed stack by $1.24\times$ offline and $1.9\times$ in model serving under a one-second p99 latency target.
For offline batch inference, we implement MIS-based scoring as a reusable, model-agnostic Ray Data API, proposed upstream, and model serving uses MIS support contributed to SGLang.
\end{abstract}

]

\printAffiliationsAndNotice{\mlsysEqualContribution}

\section{Introduction}
\label{sec:intro}
 Large language models (LLMs) are increasingly used to label, classify, and judge content \citep{gilardi2023chatgpt,zheng2023judging}.
A common industrial pattern precomputes which candidate labels apply to each entity and serves the decisions from storage, so labeling cost depends on the number of entity--label judgments and on execution efficiency.

At LinkedIn, \sysname identifies which qualifications from a catalog of skills, experience, education, and licenses are supported by the information in a member profile.
Labeling runs in offline batch inference and on a streaming update path for profile changes, and downstream matching and ranking reuse the stored predictions.
Each prediction should therefore depend only on its profile--qualification pair.
The catalog and downstream models are outside the paper's scope.

Each member's workload pairs one long profile with many short candidate qualifications supplied by retrieval.
The existing LLM labeler uses \emph{packed generation}, a form of batch prompting \citep{cheng2023batchprompting}, to answer several qualifications in one prompt, sharing profile encoding but making each answer depend on the other qualifications and on earlier answers (Section~\ref{subsec:baseline}).
The key distinction is between sharing computation and coupling predictions.
Packing shares the profile encoding, but generating independent decisions as one answer list also couples them, spends decoding on the list, and relies on the list being well-formed.

We instead use an end-to-end recipe that reformulates labeling as \emph{semantic scoring}, which treats each judgment as pointwise ranking and assigns an independent score to each profile--qualification pair from one prefill pass (Section~\ref{sec:scoring}).
Scoring one candidate per request keeps decisions independent but repeats the profile in every request, sustaining only about 90 labels/s in model serving against 2.5K for packed generation (Section~\ref{subsec:eval-ablation}).
Multiple-Item Scoring (MIS), which our prior work introduced for online LLM ranking \citep{li2026mixlm,borisyuk2026semanticsearch}, makes the formulation practical by executing a round of these scores in one request, sharing profile encoding without cross-item interaction (Section~\ref{sec:mis}).

Semantic scoring changes at most 0.01\% of labels when candidates are reordered, at least $86\times$ fewer than packed generation with the most training data (Section~\ref{subsec:eval-quality}).
Its agreement with reference labels at 45K training pairs takes packed generation about 210K pairs to match.
Compared with the deployed packed-generation stack, our MIS-based scoring stack raises labeling throughput by $1.24\times$ in offline batch inference at each stack's best batch configuration and by $1.9\times$ in model serving for profile updates at a matched one-second 99th-percentile latency target (Section~\ref{subsec:eval-throughput}).
These are stack-level comparisons: the stacks also differ in engine, checkpoint, prompt format, and runtime tuning, so the ratios measure the integrated system rather than MIS alone.

Answer-token readout, label-only fine-tuning, and MIS with its candidate-isolating mask each have precedents (Section~\ref{sec:related}); MIS-based scoring as a batch-inference API in Ray Data is new.
We contribute a training and inference system that moves a production labeler from generation to scoring:
\begin{enumerate}
  \item \textbf{Generation-to-scoring recipe.} We replace generated answer lists with independent per-item scores through shared-prefix prompts, label-only fine-tuning on reference labels, and an answer-token readout whose decision threshold can be set without retraining (Section~\ref{sec:scoring}).
  \item \textbf{MIS scoring for batch inference.} We adapt MIS \citep{li2026mixlm} from online ranking to labeling and analyze its cost in prefill tokens per label (Section~\ref{sec:mis}). We implement it as a model-agnostic scoring API for Ray Data batch inference, proposed upstream, which turns one shared query and its candidates into item-aligned scores (Section~\ref{sec:serving}).
  \item \textbf{Deployment and evaluation on both paths.} We run one per-candidate decision contract in both offline batch inference and model serving for profile updates, with MIS in SGLang for model serving (Section~\ref{sec:serving}). We evaluate agreement with reference labels, order sensitivity, stack-level throughput, and request sizing for one long profile with hundreds of short candidates (Sections~\ref{sec:evaluation}--\ref{sec:production}).
\end{enumerate}

\section{Background and Problem Setting}
\label{sec:background}

\begin{figure}[t]
  \centering
  \includegraphics[width=\columnwidth]{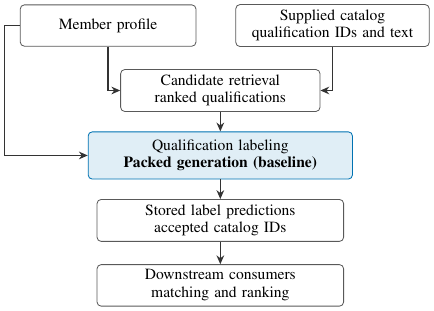}
  \caption{Profile qualification labeling against a supplied catalog.
  The baseline uses packed generation for the highlighted labeling stage.
  Sections~\ref{sec:scoring} and~\ref{sec:mis} replace that stage with semantic scoring and MIS.
  Stored predictions are reused by downstream consumers.}
  \label{fig:architecture}
\end{figure}

\subsection{Case study: profile qualification labeling}
\label{subsec:labels}
 \Sysname takes a supplied catalog of qualifications describing skills, experience, education, or licenses.
Each entry has a canonical ID, description, category, and embedding.
For example, the synthetic profile ``Registered nurse; five years in intensive care'' supports the qualifications ``holds a registered nurse license'' and ``has intensive-care nursing experience.''

The labeler is fine-tuned on profile--qualification pairs with reference \textsc{yes}/\textsc{no} labels.
How these reference labels are produced, and how the retriever is trained, are outside the scope of this paper and unchanged by this work.

The \emph{retrieve-then-label} process embeds a profile, retrieves a ranked list of $K$ catalog qualifications, and labels consecutive \emph{rounds} of $N$ candidates (Figure~\ref{fig:architecture}).
Here $K$ is in the thousands and $N$ in the hundreds.
The labeler evaluates at most $K$ qualifications per member in at most $\lceil K/N \rceil$ rounds.
After each round, labeling stops if the accepted fraction falls below a threshold.
The rule assumes that lower-ranked qualifications are less likely to be accepted by the labeler.
Both retrieval and early stopping trade recall for cost: qualifications outside the top $K$ are never judged, and stopping can miss positive labels later in the list.
Accepted IDs form the profile's stored label set, which downstream systems reuse without rerunning the labeler.
These labels are model predictions from profile information, not independently verified qualifications.
Stored labels are one input to downstream systems, and this paper does not evaluate downstream decisions.
Profile changes trigger a streaming update path that recomputes the profile's full label set.
Offline batch inference processes recent updates and periodically refreshes labels.
Semantic scoring and MIS replace the labeling stage without changing retrieval, early stopping, or the stored-label interface.

\subsection{Workload shape and notation}
\label{subsec:workload}
Each judgment combines a \emph{query} $q$, which holds the task instruction, member profile, and other candidate-invariant text, with an \emph{item} $c_i$.
The item contains the $i$-th candidate qualification $\ell_i$ and the template suffix ending immediately before the model's \textsc{yes} or \textsc{no} answer.
We write $L_q = |q|$ and $L_i = |c_i|$ in tokens.
Across the benchmark workload, the mean query is about 27 times as long as the mean item ($\rho_{\mathrm{work}} \approx 27$, Section~\ref{subsec:eval-setup}).
In \emph{independent scoring}, each query--item pair forms a separate request.
Without prefix reuse, most input tokens are copies of the shared query.

\subsection{System requirements}
\label{subsec:requirements}
\paragraph{Quality.}
The semantic-scoring labeler should reach label quality comparable to the packed-generation baseline's.
We measure label quality as agreement with reference labels (RQ1, Section~\ref{subsec:eval-quality}).
\paragraph{Consistency.}
For a fixed model and profile--\hspace{0pt}qualification pair, the judgment must be consistent across offline batch inference and the streaming update path, regardless of the other candidates in the prompt or their order.
We measure decision flips under candidate reordering (RQ1).
\paragraph{Throughput and cost.}
Offline throughput determines how quickly stored labels can be refreshed after a model or catalog update and how much GPU time each refresh consumes (RQ2, RQ3).
\paragraph{Freshness.}
The streaming update path must absorb bursts of profile edits within a latency target so members' labels can be updated without waiting for the next batch refresh.
We evaluate it as throughput at a fixed tail-latency target (RQ2), not as end-to-end label freshness.

\subsection{Packed-generation baseline and its trade-off}
\label{subsec:baseline}
Within each round (Section~\ref{subsec:labels}), the packed-generation baseline divides the $N$ candidates into $\lceil N/g \rceil$ prompts of up to $g$ candidates, each sent as a separate request.
Each full prompt contains $q$ and an ordered list $\ell_{1:g}$ of qualifications, and a fine-tuned LLM generates a comma-separated answer list such as \textsc{yes}, \textsc{no}, \textsc{yes}.
The $i$-th answer labels $\ell_i$.
Let $o=(o_1,\ldots,o_T)$ be the $T$ generated tokens of the serialized answer list, including commas and the end-of-sequence token.

For model parameters $\theta$, the probability of the answer list factorizes autoregressively as
\begin{equation}
  P_\theta(o \mid q, \ell_{1:g}) = \prod_{t=1}^{T} P_\theta(o_t \mid q, \ell_{1:g}, o_{<t}),
  \label{eq:generation}
\end{equation}
where $o_{<t}=(o_1,\ldots,o_{t-1})$ is the prefix before token $t$.
Packing encodes the profile once per prompt and spreads request overhead across $g$ decisions, but generating the answers jointly involves three design trade-offs:
\begin{enumerate}
  \item \textbf{Prompt dependence.}
  Each answer is conditioned on every qualification in the prompt and on earlier answers (Equation~\eqref{eq:generation}), so a judgment can vary with its companions, its position, or earlier answers.
  Section~\ref{sec:evaluation} measures decision flips under reordering.
  \citet{cheng2023batchprompting} likewise report that the number of samples per prompt affects accuracy.
  Training on shuffled lists of varying length exposes the model to different counts and orders but does not enforce invariant answers.
  \item \textbf{Decoding and output format.}
  The $T$ answer-list tokens are generated autoregressively, then parsed and mapped to qualifications by position, so labeling relies on the expected list format.
  The model is fine-tuned to produce this format, so its loss supervises both label decisions and answer-list formatting.
  \item \textbf{Fixed decisions.}
  The labeler returns hard decisions rather than scores that can be thresholded to trade precision for recall.
  Answer-token probabilities could be read during decoding, but each is conditioned on earlier answers, so it is not an independent per-item score.
\end{enumerate}

Independent semantic scoring, defined in Section~\ref{sec:scoring}, evaluates each profile--qualification pair in its own request.
This preserves independence but requires $N$ requests per round, each containing the same query.
Prefix caching \citep{kwon2023vllm,zheng2024sglang} reduces this recomputation only when requests reach a worker still holding the query's cache (Section~\ref{subsec:mis-cost}).

The surrounding workflow already resembles a ranking pipeline: retrieval supplies an ordered candidate list, and early stopping controls how much of that list is judged.
The recipe keeps that pipeline and recasts the labeler as a pointwise ranker with $q$ as the query and each $c_i$ as an item, changing one part of the labeler per stage (Table~\ref{tab:recipe}).

\begin{table}[t]
  \caption{Recipe stages: the change each makes to the packed-generation labeler and the check the recipe calls for before the next stage.}
  \label{tab:recipe}
  \centering
  \footnotesize
  \setlength{\tabcolsep}{3pt}
  \begin{tabular}{@{}p{0.17\columnwidth}p{0.37\columnwidth}p{0.41\columnwidth}@{}}
    \toprule
    \raggedright Stage & \raggedright Change & \raggedright Check before the next stage \tabularnewline
    \midrule
    \raggedright Prompt (\S\ref{subsec:scoring-prompt}) & \raggedright One decision per prompt, shared text in $q$ & \raggedright Single-token answers, split tokens match \tabularnewline
    \raggedright Training (\S\ref{subsec:scoring-training}) & \raggedright Reference label per item, label-only loss & \raggedright Held-out quality on the served prompt \tabularnewline
    \raggedright Readout (\S\ref{subsec:scoring-rule}) & \raggedright Score from \textsc{yes}/\textsc{no} logits, tuned threshold $\tau$ & \raggedright Held-out FP/FN against references at chosen $\tau$ \tabularnewline
    \raggedright Inference (\S\ref{sec:mis}) & \raggedright MIS scores a round in one request & \raggedright Matches single-item scores, $N$ fits limits \tabularnewline
    \raggedright Integration (\S\ref{sec:serving}) & \raggedright Only the labeler changes, on both paths & \raggedright Same checkpoint, answer tokens, and $\tau$ on all paths \tabularnewline
    \bottomrule
  \end{tabular}
\end{table}

\section{From Generation to Semantic Scoring}
\label{sec:scoring}
 The first three stages of the recipe define the prompt, train the labeler, and convert its answer-token logits into a score.
In our case study, each input pairs a profile with one qualification, and a reference answer, used only for training and validation, indicates whether the profile supports it.
\emph{Semantic scoring} assigns one score to each query--item pair $[q; c_i]$ (Section~\ref{subsec:workload}), and MIS later computes many such scores together without changing them.

\subsection{Prompt design}
\label{subsec:scoring-prompt}
\paragraph{Decision contract.}
The prompt asks for a decision about a single candidate and states what counts as supporting evidence and how to treat missing or ambiguous information.

\paragraph{Shared-prefix layout.}
Instructions, shared context, and other candidate-invariant text belong in $q$, while $c_i$ contains the candidate and the suffix leading to the answer position.
For example, the field marker \textsf{Qualification:} can end $q$ instead of repeating in every item, but candidate-dependent text must stay in $c_i$, and moving this boundary requires revalidation.

\paragraph{Answer position.}
The rendered prompt asks for a single answer, and a compact format removes the chat-template trailer that would otherwise repeat in every item, as in this synthetic example:
\begin{center}
  \footnotesize
  \setlength{\tabcolsep}{2pt}
  \begin{tabular}{@{}ll@{}}
    \toprule
    Prompt & Query $q$ $\Vert$ item $c_i$ \\
    \midrule
    Chat & \textsf{Profile: \ldots\ $\Vert$ Qualification: ICU nurse $\langle$eot$\rangle\langle$asst$\rangle$} \\
    Compact & \textsf{Profile: \ldots\ Qualification: $\Vert$ ICU nurse $\langle$r$\rangle$} \\
    \bottomrule
  \end{tabular}
\end{center}
Here $\langle$eot$\rangle\langle$asst$\rangle$ closes the user turn and opens the assistant turn, and $\langle$r$\rangle$ is a single readout token, an existing special token of the base model.
The compact prompt shortens items from about 28 to about 24 tokens on average.
The readout needs single-token answer labels in the rendered answer context, and tokenizing the query and item separately must yield the same tokens as the full single-item prompt.

\subsection{Model training}
\label{subsec:scoring-training}
\paragraph{Reference data.}
Training regroups the reference-labeled pairs used for packed generation into one qualification per example (Section~\ref{subsec:labels}).
Each example $(q, c_i, a_i)$ carries a reference answer token $a_i \in \{\textsc{yes}, \textsc{no}\}$, with $a_i = \textsc{yes}$ when the reference judgment indicates that the profile information supports $\ell_i$.
The decision semantics are unchanged, so the labeler is trained on these hard labels through supervised fine-tuning (SFT).
Members are split before pairing, so the member-disjoint holdout is unchanged.

\paragraph{Model and objective.}
Fine-tuning starts from the base instruction-tuned model and minimizes the answer label's negative log-likelihood, $-\log P_\theta(a_i \mid q, c_i)$, masking all other tokens.
This standard causal-LM loss over the full vocabulary concentrates supervision on the qualification decision, and inference later renormalizes over the two answer tokens (Section~\ref{subsec:scoring-rule}).
Single-item training removes answer-list length and formatting from the prediction target.
Training does not use MIS: MIS is applied only at inference, in validation and on both labeling paths, where its masking keeps each packed item isolated as in single-item training (Section~\ref{sec:mis}).
Model size is then chosen by held-out quality at the desired operating point and by cost (Section~\ref{subsec:eval-small-model}).

\paragraph{Prompt--checkpoint contract.}
Training, validation, and serving use the same prompt renderer, answer-token mapping, and answer position, so a checkpoint is validated in the format it is served in.
A compact prompt or a different readout token therefore needs matched fine-tuning and validation.

\subsection{Score readout and validation}
\label{subsec:scoring-rule}
The single-item prompt $[q; c_i]$ ends immediately before the answer, so a prefill pass provides the next-token logits $z_i$ at the item's last token ($\langle$asst$\rangle$ in the chat prompt, $\langle$r$\rangle$ in the compact prompt) without decoding.
Renormalizing over the \textsc{yes} and \textsc{no} answer tokens gives the score
\begin{equation}
  s_i = \frac{\exp z_i(\textsc{yes})}{\exp z_i(\textsc{yes}) + \exp z_i(\textsc{no})},
  \label{eq:score}
\end{equation}
and qualification $\ell_i$ is accepted when $s_i \ge \tau$.
The default $\tau = 0.5$ selects the more likely answer, with ties assigned to \textsc{yes}.
Tuning $\tau$ on validation data selects the precision--recall operating point without retraining the labeler.
Retained scores support calibration analysis and inspection of borderline cases, but two-token renormalization does not calibrate them.
Each score depends only on $[q; c_i]$, so it is invariant to other candidates and their order.

Before enabling MIS, the recipe fixes $\tau$ on validation data and compares held-out agreement with reference labels against the packed-generation baseline on the same reference set (Section~\ref{subsec:eval-quality}).

\section{Inference with Multiple-Item Scoring}
\label{sec:mis}

\subsection{Mechanism}
\label{subsec:mis-mechanism}
 In the recipe's inference stage, MIS computes all $N$ scores of a round in one prefill pass while keeping each score $s_i$ dependent only on $[q; c_i]$ (Section~\ref{sec:scoring}).
We adapt MIS, introduced for online LLM ranking \citep{li2026mixlm,borisyuk2026semanticsearch} and contributed by us to FlashInfer and SGLang \citep{ye2025flashinfer,zheng2024sglang}, to semantic labeling.
Each $[q; c_i]$ uses the single-item prompt of Section~\ref{subsec:scoring-prompt}.
The engine prefills one physical token sequence
\begin{equation}
  x = [\,q;\, c_1;\, c_2;\, \ldots;\, c_N\,],
  \label{eq:mis-packing}
\end{equation}
with segment delimiters omitted from the notation.
Segment boundaries come from the item lengths, which set the attention mask and position IDs.
Let $j$ and $k$ be token indices in the packed sequence $x$.
Write $\mathcal{Q}$ for the set of indices occupied by $q$ and $\mathcal{I}_i$ for those occupied by $c_i$.
These sets are contiguous and disjoint.

Position IDs encode token positions for the model, and unlike packed-sequence indices, they can be reused across items.
Query tokens receive position IDs $0, \ldots, L_q - 1$, and the token at zero-based offset $d$ in every item receives $L_q + d$, as in its single-item prompt.
Every item therefore starts at position ID $L_q$, as if it immediately followed $q$ in a separate request.
Recovering independent scoring requires both the attention mask and these reused position IDs, because later items would otherwise have different positions relative to the query.

\begin{figure}[!t]
  \centering
  \includegraphics[width=0.9\columnwidth]{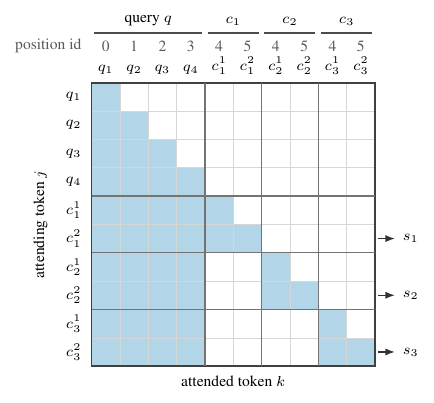}
  \caption{MIS attention mask for a synthetic packed sequence $x = [q; c_1; c_2; c_3]$: each item token attends only to $q$ and its own causal prefix, and every item reuses the position IDs that follow $q$.}
  \label{fig:mis-mask}
\end{figure}

Attention is restricted by the additive mask
\begin{equation}
  M_{jk} =
  \begin{cases}
    0 & \text{if } k \le j \text{ and } k \in \mathcal{Q},\\
    0 & \text{if } k \le j \text{ and } j,k \in \mathcal{I}_i,\\
    -\infty & \text{otherwise.}
  \end{cases}
  \label{eq:mis-mask}
\end{equation}
The first case lets every token, in $q$ or in any item, attend causally to the query tokens, so all items share $q$; the second lets item tokens attend causally within their own item $c_i$.
Each item attends only to $q$ and its own causal prefix (Figure~\ref{fig:mis-mask}).
Together, the mask and reused position IDs give the final token of $c_i$ the same context as in $[q; c_i]$.
Equation~\eqref{eq:score} yields the independent score $s_i$ up to floating-point differences, regardless of other items or their order; such differences can still flip a thresholded decision whose score lies very close to $\tau$.
Section~\ref{subsec:eval-quality} tests the resulting invariance to candidate order.

\subsection{Cost of scoring one query}
\label{subsec:mis-cost}
Table~\ref{tab:mis-cost} compares the work required to score the $N$ items of a round for one query.
Prefix caching can reuse the query only while its cache entries remain resident on the worker handling each request.
Even with perfect reuse, the engine must schedule $N$ sequences, maintain their metadata, and collect $N$ outputs.
MIS consolidates that work into one sequence with all $N$ scores in one output.
Generating a single answer token would also need only one prefill pass, so the decode savings in Table~\ref{tab:mis-cost} come from removing the baseline's multi-token answer list.

\begin{table}[t]
  \caption{Work to score the $N$ items of one query ($g$ items per packed-generation prompt; delimiters and template tokens omitted).}
  \label{tab:mis-cost}
  \centering
  \footnotesize
  \setlength{\tabcolsep}{2.3pt}
  \begin{tabular}{@{}lcccc@{}}
    \toprule
    & Packed gen. & Independent & + prefix cache & MIS \\
    \midrule
    Requests & $\lceil N/g \rceil$ & $N$ & $N$ & $1$ \\
    Query prefill & $\lceil N/g \rceil L_q$ & $N L_q$ & $L_q$ to $N L_q$ & $L_q$ \\
    Decoded tokens & $\Theta(N)$ & $0$ & $0$ & $0$ \\
    Cross-item attn. & yes & no & no & no \\
    Output & text & score & score & score \\
    \bottomrule
  \end{tabular}
\end{table}

\subsection{Cost model}
\label{subsec:mis-cost-model}
We measure cost in prefill tokens per label, because MIS decodes nothing and long prefill-only requests saturate the GPU at the latency knee (Section~\ref{subsec:eval-throughput}).
On the benchmark (Section~\ref{subsec:eval-setup}), independent requests without reuse prefill about 770 tokens per label, packed generation about 63, before reuse of its cached instruction prefix, plus up to two decoded tokens, and MIS at $N = 200$ about 32.
Packed generation thus takes about $2.0\times$ the input tokens per label of MIS, because MIS prefills the query once per round instead of once per prompt, and MIS also removes answer-list decoding.
The measured $1.9\times$ model-serving speedup is of similar size (Section~\ref{subsec:eval-throughput}), but throughput also depends on cache hits, batching, host-side work, and the engine, so the token ratio does not by itself explain it.

More generally, for a request with $N$ items, let $\bar{L}_{\mathrm{req}} = \frac{1}{N}\sum_{i=1}^{N} L_i$ be the mean item length and $\rho_{\mathrm{req}} = L_q/\bar{L}_{\mathrm{req}}$ the query-to-item length ratio.
Then the prefill-token reduction factor relative to independent scoring without prefix reuse is
\begin{equation}
  R(N) = \frac{N L_q + \sum_{i} L_i}{L_q + \sum_{i} L_i} = \frac{N(1+\rho_{\mathrm{req}})}{N+\rho_{\mathrm{req}}},
  \label{eq:mis-reduction}
\end{equation}
which approaches $1+\rho_{\mathrm{req}}$ as $N$ grows.
$R(N)$ is an exact ratio of prefill tokens, and compute follows it only up to constant factors.
Feed-forward (MLP) and projection work scale with this token count.
For items no longer than the query, attention work is $O(L_q^2 + N L_q \bar{L}_{\mathrm{req}})$ rather than $O\bigl((L_q + N \bar{L}_{\mathrm{req}})^2\bigr)$ for a fully causal packed sequence, because MIS computes the query's self-attention once and each item attends only to $q$ and itself, and the FlashInfer kernels skip fully masked tiles.
Section~\ref{subsec:eval-ablation} compares $R(N)$ with measured speedups.

\subsection{Request sizing}
\label{subsec:mis-sizing}
Each item has the logical context $[q; c_i]$, and the largest position ID in a packed request is $L_q + \max_i L_i - 1$.
The longest single-item prompt therefore determines the positional span.
The physical MIS request still contains $L_q + \sum_i L_i$ tokens, so it must satisfy the engine's aggregate input-length limit and the request's key--value (KV) cache budget.
Splitting items across requests reduces aggregate length, but each split request prefills $q$ again unless prefix caching reuses it.
Large requests also leave less room for batching, which matters most for model serving for profile updates.
In production, each retrieval round of Section~\ref{subsec:labels} becomes one MIS request, so $N$ is the round size.
The benchmark in Section~\ref{sec:evaluation} uses $N = 200$ so that one MIS request scores the same candidates per member as packed generation's ten prompts of $g = 20$.

\section{Batch and Model-Serving Integration}
\label{sec:serving}
\begin{figure}[!t]
  \centering
  \includegraphics{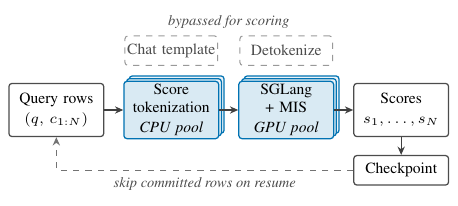}\\[-1pt]
  {\footnotesize (a) Offline batch pipeline}\\[4pt]
  \includegraphics{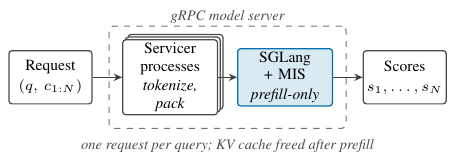}\\[-1pt]
  {\footnotesize (b) Model serving for profile updates}
  \caption{MIS changes only the scoring stage of both serving paths.
  (a)~Offline: tokenization and scoring (blue) run in separate autoscaling actor pools and bypass the generation-only stages (dashed).
  (b)~Profile updates: a multi-process gRPC server \citep{li2026mixlm} sends each packed request to a prefill-only SGLang engine.}
  \label{fig:mis-serving}
\end{figure}

 In the case study, MIS replaces only the labeler in offline batch inference and the streaming update path (Figure~\ref{fig:mis-serving}).
Retrieval, rounds, early stopping, and the stored-label schema are unchanged (Section~\ref{subsec:labels}), so downstream consumers need no interface changes.
Each round becomes one request carrying the profile as the query and the $N$ candidate IDs with their item text.
It returns one score per candidate in the same order, thresholded as in Section~\ref{subsec:scoring-rule}.
Both paths share the checkpoint, answer tokens, and threshold.
The item layout is set per path: the offline benchmark uses the compact layout and model serving uses the chat layout (Section~\ref{subsec:scoring-prompt}), so cross-path agreement depends on validating each layout.

\paragraph{Offline.}
A scheduled workflow prepares one input row per member and round, then scores it in a Ray Data pipeline \citep{moritz2018ray} (Figure~\ref{fig:mis-serving}(a)).
Separate autoscaling actor pools let CPU tokenization and GPU inference with SGLang scale independently.
Ray Data's open-source SGLang integration supported only text generation, which maps each input row to one prompt and one completion.
We implemented a model-agnostic MIS scoring API for it, usable with any MIS-compatible SGLang model and proposed upstream, that maps each row to one shared query and its items, tokenizes them as separate segments, and returns one score per item in input order.
Inputs already include the chat template, so the API skips further templating and detokenization.
Completed rows are persisted, so an interrupted job resumes where it stopped.
Round-level early stopping follows Section~\ref{subsec:labels}, with only eligible members proceeding to the next round.
Accepted labels are written to a key-value store for downstream reuse.

\paragraph{Profile updates.}
Profile updates trigger a service that embeds the profile, retrieves candidates, and sends each round as a request to a prefill-only SGLang model server (Figure~\ref{fig:mis-serving}(b)), whose MIS support is available in open-source SGLang \citep{zheng2024sglang}.
The server reuses components from prior LLM ranking systems \citep{behdin2025sjs,li2026mixlm}.
Servicer processes tokenize and pack each request with the mask in Equation~\eqref{eq:mis-mask}, keeping request handling off the engine's event loop.
The engine returns item-aligned scores without sampling or decoding.
It collects a request's scores in one GPU-side gather, avoiding many small device-to-host copies, and frees the request's KV cache once prefill completes.
Because each round of candidates is processed as one sequence, the scheduler needs neither coordinated batching of separate item requests nor in-batch prefix caching to share the profile \citep{behdin2025sjs}.

\section{Evaluation}
\label{sec:evaluation}
 We evaluate the recipe on the case study through decision quality and efficiency:
\textbf{RQ1:} How does semantic scoring compare with packed generation in agreement with reference labels, consistency under reordering, and training-data needs?
\textbf{RQ2:} What throughput do the deployed generation and scoring stacks achieve under matched workload and latency requirements?
\textbf{RQ3:} How does MIS throughput vary with request size?
\textbf{RQ4:} (Exploratory) How much further can a smaller scoring labeler raise throughput?

\subsection{Setup}
\label{subsec:eval-setup}
\paragraph{Workload.}
The throughput benchmark contains approximately 500K member profiles, each with 200 retrieved candidate qualifications, for a total of approximately 100M query--item pairs.
The query averages 744 tokens (median 543, 95th percentile 2{,}047) and items average 28 tokens, so the query-to-item ratio of workload mean lengths is $\rho_{\mathrm{work}} \approx 27$.
Query lengths vary by more than an order of magnitude across members, whereas item lengths are nearly constant.
Label quality is measured against reference judgments (Section~\ref{subsec:labels}) on a held-out set split by member, so no member contributes pairs to both training and evaluation.
Both the training and held-out sets are fairly balanced, with a positive rate of about 50\%.
The metric thus measures agreement with these references, and catalog gaps, retrieval misses, qualifications skipped by early stopping, and product-level outcomes fall outside this evaluation (Section~\ref{sec:discussion}).

\paragraph{Methods.}
\emph{Packed generation} is the production baseline from Section~\ref{subsec:baseline}, with $g = 20$ qualifications per prompt, about 1{,}300 tokens per prompt, and up to 40 greedily decoded tokens.
\emph{Independent semantic scoring} places one qualification in each request.
It appears as the $N = 1$ point of RQ3, with cross-request prefix caching disabled.
\emph{MIS} packs $N$ qualifications per request.
With the chat-template item layout, a member's request at $N = 200$ holds about 6{,}400 tokens, against 154{,}000 for independent requests without reuse.
Using $\rho_{\mathrm{work}}$ in Equation~\eqref{eq:mis-reduction} gives the workload-level estimate $R(200) \approx 24$.
Unless noted, all methods run internally fine-tuned 1.5B-parameter decoder-only LLMs.
Packed generation uses its answer-list fine-tune, and the scoring methods share the single-item fine-tune.

\paragraph{Hardware and engines.}
Offline runs use one NVIDIA H100 GPU in the pipeline of Section~\ref{sec:serving}, and model-serving benchmarks for profile updates use one NVIDIA H200 GPU, matching the production serving hardware.
As in production, packed generation runs on vLLM \citep{kwon2023vllm} with chunked prefill and prefix caching.
Independent scoring and MIS run on SGLang with FlashInfer kernels \citep{zheng2024sglang,ye2025flashinfer}, because MIS is supported in SGLang but not in vLLM.

\begin{figure}[!t]
  \centering
  \includegraphics{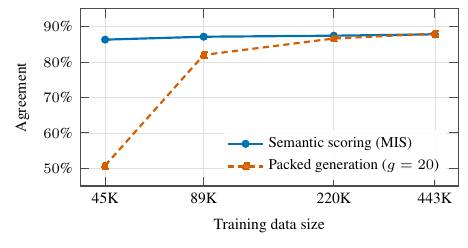}\par
  \caption{Agreement with reference labels against training-data size (profile--qualification pairs) for semantic scoring and packed generation.}
  \label{fig:eval-accuracy}
  \vspace{0.5em}
  \includegraphics{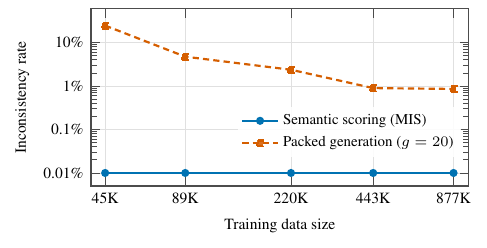}
  \caption{Inconsistency rate (share of labels that change when a member's qualifications are reordered and regrouped, log scale) against training-data size.}
  \label{fig:eval-consistency}
\end{figure}
\subsection{RQ1: Quality}
\label{subsec:eval-quality}
We evaluate output quality as agreement with the reference labels and consistency when qualifications are reordered.
Semantic scoring agrees well with the references with little training data and is consistent at every training size, whereas packed generation needs several times more data to match its agreement and never matches its consistency.
Semantic-scoring checkpoints are trained pointwise and evaluated with MIS inference.

\paragraph{Agreement with reference labels.}
Agreement is the percentage of predicted qualification labels that match the reference labels of Section~\ref{subsec:eval-setup}.
We train semantic-scoring and packed-generation labelers at several training-data sizes, and Figure~\ref{fig:eval-accuracy} plots their agreement.
Semantic scoring reaches 86.3\% with 45K training pairs.
Packed generation improves more slowly: at 45K pairs it reaches 50.8\%, near the 50\% positive rate of this balanced set, and it needs about 210K pairs, interpolating between measured sizes, or about $4.7\times$ as much data, to match semantic scoring's agreement at 45K.
At the largest size with agreement results, 443K pairs, the two are comparable, at 88.0\% and 87.8\%; we make no significance claim for this difference.

\paragraph{Consistency.}
Consistency rate is measured as the percentage of qualification labels that stay the same when a member's qualifications are randomly reordered and regrouped during evaluation, with the profile and qualification text held fixed.
Figure~\ref{fig:eval-consistency} plots the inconsistency rate, the share of labels that change across orderings, for the same training sizes plus an 877K set, for which agreement is not reported.
Semantic scoring keeps it at or below 0.01\% at every size; the mechanism predicts no change, so this residual reflects floating-point differences in batched execution (Section~\ref{subsec:mis-mechanism}).
Packed generation changes 24.8\% of labels at the smallest training size and, although it improves with more data, still changes 0.86\% at the largest, at least $86\times$ the rate of semantic scoring.

\paragraph{A possible explanation.}
Both formulations are fine-tuned on the same reference-labeled pairs and differ only in how pairs are grouped into training examples (Section~\ref{subsec:scoring-training}).
A packed-generation example supervises both the individual judgments and the output format: the model must emit a complete answer list whose length and order match its candidates.
Semantic scoring supervises a single answer token per pair, so list length, formatting, and alignment never enter the target, and each decision is independent of candidate order by construction.
This is consistent with both gaps above: packed generation likely spends its early training budget on the list format, leaving agreement near the positive rate at the smallest size, and its decisions remain sensitive to candidate position at every size.
Semantic scoring has a training-cost trade-off: single-item examples repeat the profile for every pair, so training processes more tokens per labeled pair, and MIS recovers this sharing only at inference.

\subsection{RQ2: Throughput}
\label{subsec:eval-throughput}
\paragraph{Protocol.}
Throughput is measured in labels, i.e., scored query--item pairs, per second per GPU on the same candidate workload.
Offline, each method runs at its best batch size and concurrency.
For model serving for profile updates, every method is measured at the knee: the highest Poisson-arrival request rate that holds the 99th-percentile end-to-end request latency at or below one second with at least 99\% success.
Each rate runs for 60 seconds after a 200-request warmup on disjoint client and server CPU sets, with the model server restarted so every run begins with an empty prefix cache.
Each member's ten packed-generation prompts are separate requests that may reach different hosts, so fan-out limits cache reuse to the shared instruction prefix.
MIS issues one request per member and is insensitive to routing.
\begin{table}[!tb]
  \caption{Throughput of the deployed inference stacks, in labels per second per GPU, with the ratio to packed generation in parentheses.
  Offline (H100): steady state at the best batch size and concurrency; profile updates (H200): highest arrival rate with p99 latency $\le$1\,s.
  The stacks differ in engine, checkpoint, prompt format, and runtime tuning as well as method.}
  \label{tab:eval-throughput}
  \centering
  \footnotesize
  \begin{tabular}{@{}lrr@{}}
    \toprule
    Method & Offline & Profile updates \\
    \midrule
    Packed generation ($g = 20$) & 5.1K (1.0$\times$) & 2.5K (1.0$\times$) \\
    MIS ($N = 200$) & 6.3K (1.24$\times$) & 4.8K (1.9$\times$) \\
    \bottomrule
  \end{tabular}
\end{table}

\paragraph{Model serving for profile updates.}
At the one-second p99 target, the MIS-based stack at $N = 200$ sustains about 4.8K labels/s per GPU against 2.5K for packed generation, a $1.9\times$ speedup at equal latency (Table~\ref{tab:eval-throughput}).
At the knee, MIS is latency-bound and the GPU is saturated by long prefill-only requests.
The gain is consistent with removing repeated query prefill, per-request overhead, and answer-list decoding, but because the comparison changes both the formulation and the inference stack, we cannot quantify each factor's contribution.

\paragraph{Offline batch inference.}
On the 500K-member job, the MIS-based stack processes about 6.3K labels/s per GPU against 5.1K for packed generation, a $1.24\times$ throughput ratio (Table~\ref{tab:eval-throughput}).
Reaching this rate required tuning the offline pipeline, not only changing the method.
Our first MIS run, with the chat-template layout and a single engine process, reached only 4.4K labels/s, below packed generation, and left the GPU partly idle, which points to host-side work such as tokenization, packing, and result handling rather than GPU compute.
The reported run uses the compact readout-token layout of Section~\ref{subsec:scoring-prompt}, which shortens each item, and runs two SGLang engine processes that share the GPU through CUDA Multi-Process Service (MPS), so one engine's host-side work overlaps the other's GPU work.
Together with smaller per-engine batches, these changes raise offline MIS throughput by 44\%.
The offline ratio is lower, likely because packed generation amortizes decoding across large batches without a latency target, and unchanged input and output stages take part of the job time.
Both offline numbers are steady-state throughput, excluding job start-up.
The MIS-based stack thus raises per-GPU throughput in both settings under the benchmark protocol, most under a latency target.

\subsection{RQ3: Request size}
\label{subsec:eval-ablation}
\begin{figure}[!t]
  \centering
  \includegraphics{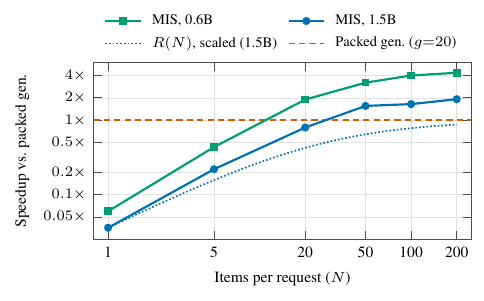}
  \caption{Model-serving speedup over 1.5B packed generation versus items per request for MIS with the 1.5B checkpoint (chat format) and the exploratory 0.6B checkpoint (readout-token format, Section~\ref{subsec:eval-small-model}), on one H200 at the one-second p99 knee without cross-request prefix caching.
  The dotted curve is $R(N)$ (Equation~\eqref{eq:mis-reduction}) scaled to the 1.5B $N = 1$ speedup.
  The two curves differ in model, checkpoint, and layout, so they do not isolate model size.}
  \label{fig:eval-ablation}
\end{figure}

RQ2 reports MIS at $N = 200$ only, and Figure~\ref{fig:eval-ablation} shows how its speedup changes from single-item scoring ($N = 1$) to larger requests, against the token-count ratio $R(N)$.

In model serving for profile updates, the sweep varies $N$ from 1 to 200 for the 1.5B checkpoint under the protocol of Section~\ref{subsec:eval-throughput}, with cross-request prefix caching disabled so that each request prefills its own profile.
Requests grow from about 770 tokens at $N = 1$ to about 6{,}400 at $N = 200$.
Throughput rises from about 90 labels/s per GPU at $N = 1$ to 2.0K at $N = 20$, 3.9K at $N = 50$, and 4.8K at $N = 200$, so MIS overtakes packed generation (2.5K) between $N = 20$ and $N = 50$.
Figure~\ref{fig:eval-ablation} also shows the exploratory 0.6B labeler (Section~\ref{subsec:eval-small-model}), which follows the same shape at higher throughput.
Splitting a member's 200 candidates into two $N = 100$ requests lowers throughput by about 14\%, to 4.1K labels/s, because the split repeats the profile prefill and adds requests.
The measured speedup exceeds the $R(N)$ curve scaled to its $N = 1$ point, possibly because packing also amortizes scheduling and output work across items.

\subsection{RQ4: A smaller decision model (exploratory)}
\label{subsec:eval-small-model}
Semantic scoring reads a single per-item decision instead of a generated answer list (Section~\ref{subsec:scoring-rule}), so a good decision model, rather than a strong completion model, may suffice.
We explore this with an internally fine-tuned 0.6B-parameter labeler on the readout-token layout of Section~\ref{subsec:scoring-prompt}.

In exploratory runs, its label quality was comparable to the 1.5B scoring labeler's; we do not report these metrics and make no general claim about the size--quality trade-off, so we report only throughput here.

The smaller labeler raises throughput on both paths.
For profile updates, under the protocol of RQ2, MIS with the 0.6B labeler sustains about 10.9K labels/s per GPU at $N = 200$ (Figure~\ref{fig:eval-ablation}), $4.4\times$ packed generation and $2.3\times$ the 1.5B labeler, a ratio that also reflects checkpoint and layout differences.
Offline, with the configuration of RQ2, it reaches an estimated 11.3K labels/s per GPU, $2.2\times$ packed generation, although this run's prompts did not use the model's own chat template, so it measures throughput only.
These results suggest that MIS and a smaller decision model can be combined, pending a full quality evaluation.

\section{Production Deployment and Impact}
\label{sec:production}
 Previously, packed generation ran on vLLM, using H100 GPUs for daily incremental and weekly offline runs and H200 GPUs in multiple regions for model serving for profile updates, including multilingual profiles.

\paragraph{MIS deployment.}
Offline incremental runs and periodic refreshes (e.g., after a model or catalog update) keep the same schedule and completion window.
Both paths now run the MIS-based scoring stack.
The benchmarks of Section~\ref{subsec:eval-throughput} use the production hardware, engine configuration, and pipeline, so their measured throughput implies about 20\% fewer GPU-hours for the labeling stage of each offline run.
MIS does not accelerate the retrieval, prompt-preparation, and output stages of each run, so end-to-end offline savings are smaller.
On the streaming update path, it implies about 50\% fewer provisioned GPUs at the same peak update rate and latency target, assuming capacity scales with per-GPU throughput.

\section{Related Work}
\label{sec:related}
Answer-token readout, hard-label fine-tuning, and MIS, including its candidate-isolating mask and reused position IDs, have precedents; our contributions are combining them to move a production labeler from generation to scoring on both labeling paths and implementing MIS-based scoring as a reusable Ray Data batch-inference API.
\paragraph{Catalog-grounded semantic labeling.}
Systems that understand job postings map posting text to standardized attributes through candidate generation and semantic selection.
\citet{li2020deepjob} generate taxonomy-based entity candidates and use learned ranking to select the most important standardized entities for a job.
\citet{xu2026unifiedsemantic} retrieve attribute candidates using embeddings and apply small-language-model classification to select the corresponding values, with occupation as the worked large-cardinality example.
Our profile labeler follows the same retrieve-then-label structure but assigns each candidate qualification an independently thresholded support score, so one profile can receive multiple qualifications.
Packed generation batches these decisions into one answer list, whereas we share computation with MIS while isolating predictions.

\paragraph{Pointwise and listwise judgments.}
Pointwise ranking with sequence-to-sequence models \citep{nogueira2020monot5} formulates document ranking as relevance-label prediction and uses the answer-token probabilities as relevance scores.
Semantic scoring applies this readout to binary qualification judgments with a causal language model (Section~\ref{sec:scoring}).
\citet{sun2023rankgpt}, by contrast, generate an ordering after seeing candidates together, making cross-candidate comparisons part of the task.
Batch prompting \citep{cheng2023batchprompting} packs separate samples into one prompt, whereas packed generation packs one entity's candidates, and both couple answers that qualification labeling needs to keep independent.

\paragraph{Shared-context execution.}
Our prior work introduced MIS in MixLM \citep{li2026mixlm} and used it in semantic search \citep{borisyuk2026semanticsearch}, sharing query computation across independent item scores to rank candidates for a live request.
We instead threshold scores into stored labels in offline batch inference and the streaming update path, which requires a generative labeler to become a scorer (Section~\ref{sec:scoring}).
Our recent article describes the SGLang scoring API for pointwise decision models \citep{ramachandran2026sglangdecision}.

Prefix caching \citep{kwon2023vllm,zheng2024sglang,gim2024promptcache} and shared-prefix attention kernels such as Hydragen \citep{juravsky2024hydragen}, ChunkAttention \citep{ye2024chunkattention}, and FlashInfer's cascade attention \citep{ye2025flashinfer} reuse or speed up query computation without coupling candidate judgments, but each candidate remains a separate sequence.
MIS instead shares the query inside one packed sequence, relying on candidate-isolating attention masks in FlashInfer.
Parallel Context Windows \citep{ratner2023pcw} similarly mask attention between context segments with reused positions, but to extend the context of one generated answer rather than to read out independent per-item scores.

\section{Discussion and Limitations}
\label{sec:discussion}
 \paragraph{Applicability.}
Semantic scoring with MIS applies when many candidates share one context, each decision depends only on that context and its candidate, and one answer position yields the score.
Candidate tasks, none evaluated here, include taxonomy tagging, checking content against independent policies \citep{inan2023llamaguard}, search relevance judgments \citep{thomas2024searcher}, and skill tagging of course transcripts \citep{yan2023skills}.
The recipe does not apply to listwise reranking \citep{sun2023rankgpt} or joint deduplication, whose decisions require the cross-candidate interactions that the MIS mask blocks.
Normalizing over single-token class labels could extend the readout to graded or multi-class decisions, though we evaluate only binary labels.

\paragraph{Limitations.}
We measure quality as agreement with reference judgments on the candidates the retriever surfaces (Sections~\ref{subsec:labels} and~\ref{subsec:eval-setup}), not as product-level outcomes.
The catalog, retriever, and early-stopping rule are unchanged by this work, so an absent label can mean either a negative prediction or a qualification that was never evaluated.
We evaluate the recipe end to end: an inconsistency rate of at most 0.01\% under reordering bounds MIS's sensitivity to packing (Section~\ref{subsec:eval-quality}), but we do not isolate per-stage contributions or numerical parity across serving paths.
Throughput comparisons are stack-level, so we do not attribute the gains to MIS alone (Section~\ref{subsec:eval-throughput}).
The smaller-model results are exploratory, and how small a scoring labeler can be without losing quality remains open (Section~\ref{subsec:eval-small-model}).

\paragraph{Deployment lessons.}
First, the prompt, the checkpoint, the answer-token mapping, and the threshold form one versioned artifact across both paths, so changing the prompt means retraining and revalidating (Table~\ref{tab:recipe}).
Second, MIS makes shared-context reuse explicit within each request, so its benefit does not depend on cache residency or request routing (Section~\ref{subsec:eval-throughput}).
Finally, request size trades amortized query work against latency and batching headroom (Section~\ref{subsec:eval-ablation}).

\section{Conclusion}
\label{sec:conclusion}
 We presented an end-to-end recipe that reformulates a generative \textsc{yes}/\textsc{no} labeler as semantic scoring through single-candidate prompts, label-only training, answer-token readout, and prefill-only execution with MIS.
In our case study, it changes at most 0.01\% of labels under reordering, matches packed generation's agreement with reference labels with about $4.7\times$ fewer training pairs, and, as a stack-level comparison, raises throughput by $1.24\times$ in offline batch inference and $1.9\times$ in model serving for profile updates.
For offline batch inference, we implement MIS-based scoring as a reusable Ray Data API, proposed upstream; model serving uses MIS in SGLang.
More broadly, industrial LLM labelers need not inherit a generative interface: when decisions are independent, the recipe separates the semantic contract, judging each candidate alone, from the systems optimization that executes many such judgments together.

\renewcommand{\bibfont}{\raggedright}
\bibliography{semantic_oracle}
\IfFileExists{mlsys2025-compact.bst}{\bibliographystyle{mlsys2025}}{%
  \IfFileExists{mlsys2026.bst}{\bibliographystyle{mlsys2026}}{%
    \IfFileExists{mlsys2025.bst}{\bibliographystyle{mlsys2025}}{%
      \bibliographystyle{plainnat}}}}

\end{document}